\documentclass[letterpaper]{article} 
\usepackage{aaai25}  
\usepackage{times}  
\usepackage{helvet}  
\usepackage{courier}  
\usepackage[hyphens]{url}  
\usepackage{graphicx} 
\usepackage{natbib}  
\usepackage{caption} 
\usepackage{algorithm}
\usepackage{algorithmic}
\usepackage{booktabs}
\usepackage{multirow}
\usepackage{amsmath}
\usepackage{xcolor}
\usepackage{dsfont}
\usepackage{sidecap}
\usepackage{amssymb}
\usepackage{colortbl}
\usepackage{indentfirst}
\usepackage{graphicx}
\usepackage{tikz}

\usepackage{newfloat}
\usepackage{listings}
\DeclareCaptionStyle{ruled}{labelfont=normalfont,labelsep=colon,strut=off} 
\floatstyle{ruled}
\newfloat{listing}{tb}{lst}{}
\floatname{listing}{Listing}
\title{DCA: Dividing and Conquering Amnesia in Incremental Object Detection}
\author{
    Aoting Zhang\textsuperscript{\rm 1,3}, Dongbao Yang\textsuperscript{\rm 1,3}\equalcontrib, Chang Liu\textsuperscript{\rm 5}, Xiaopeng Hong\textsuperscript{\rm 4}\equalcontrib, Miao Shang\textsuperscript{\rm 4}, Yu Zhou\textsuperscript{\rm 2}\\
}
\affiliations{
    \textsuperscript{\rm 1}Institute of Information Engineering, Chinese Academy of Sciences\\
    \textsuperscript{\rm 2}VCIP \& TMCC \& DISSec, College of Computer Science, Nankai University\\
    \textsuperscript{\rm 3}School of Cyber Security, University of Chinese Academy of Sciences\\
    \textsuperscript{\rm 4}Harbin Institute of Technology \\
    \textsuperscript{\rm 5}Tsinghua University\\
    \{zhangaoting, yangdongbao\}@iie.ac.cn, liuchang2022@tsinghua.edu.cn\\
    hongxiaopeng@ieee.org, miaos0522@gmail.com, yzhou@nankai.edu.cn
}

\usepackage{bibentry}

\begin{document}

\maketitle

\begin{abstract}
Incremental object detection (IOD) aims to cultivate an object detector that can continuously localize and recognize novel classes while preserving its performance on previous classes. Existing methods achieve certain success by improving knowledge distillation and exemplar replay for transformer-based detection frameworks, but the intrinsic forgetting mechanisms remain underexplored. In this paper, we dive into the cause of forgetting and discover forgetting imbalance between localization and recognition in transformer-based IOD, which means that localization is less-forgetting and can generalize to future classes, whereas catastrophic forgetting occurs primarily on recognition. Based on these insights, we propose a Divide-and-Conquer Amnesia (DCA) strategy, which redesigns the transformer-based IOD into a localization-then-recognition process. DCA can well maintain and transfer the localization ability, leaving decoupled fragile recognition to be specially conquered. To reduce feature drift in recognition, we leverage semantic knowledge encoded in pre-trained language models to anchor class representations within a unified feature space across incremental tasks. This involves designing a duplex classifier fusion and embedding class semantic features into the recognition decoding process in the form of queries. Extensive experiments validate that our approach achieves state-of-the-art performance, especially for long-term incremental scenarios. For example, under the four-step setting on MS-COCO, our DCA strategy significantly improves the final AP by 6.9\%.
\end{abstract}

%

\section{Introduction}
\noindent Object detection has experienced great advancements in recent years~\cite{he2017mask, carion2020end}. Nevertheless, prevailing models rely on fixed data, rendering them inadequate to adapt to dynamic data in real-world scenarios. To equip object detection with the ability of lifelong learning and knowledge integration like humans, incremental object detection (IOD) is proposed, which can continuously localize and recognize object instances of new concepts while maintaining old knowledge. It is a pivotal stride toward attaining artificial general intelligence.  

\begin{figure}[t]
\centering
\includegraphics[width=0.99\columnwidth]{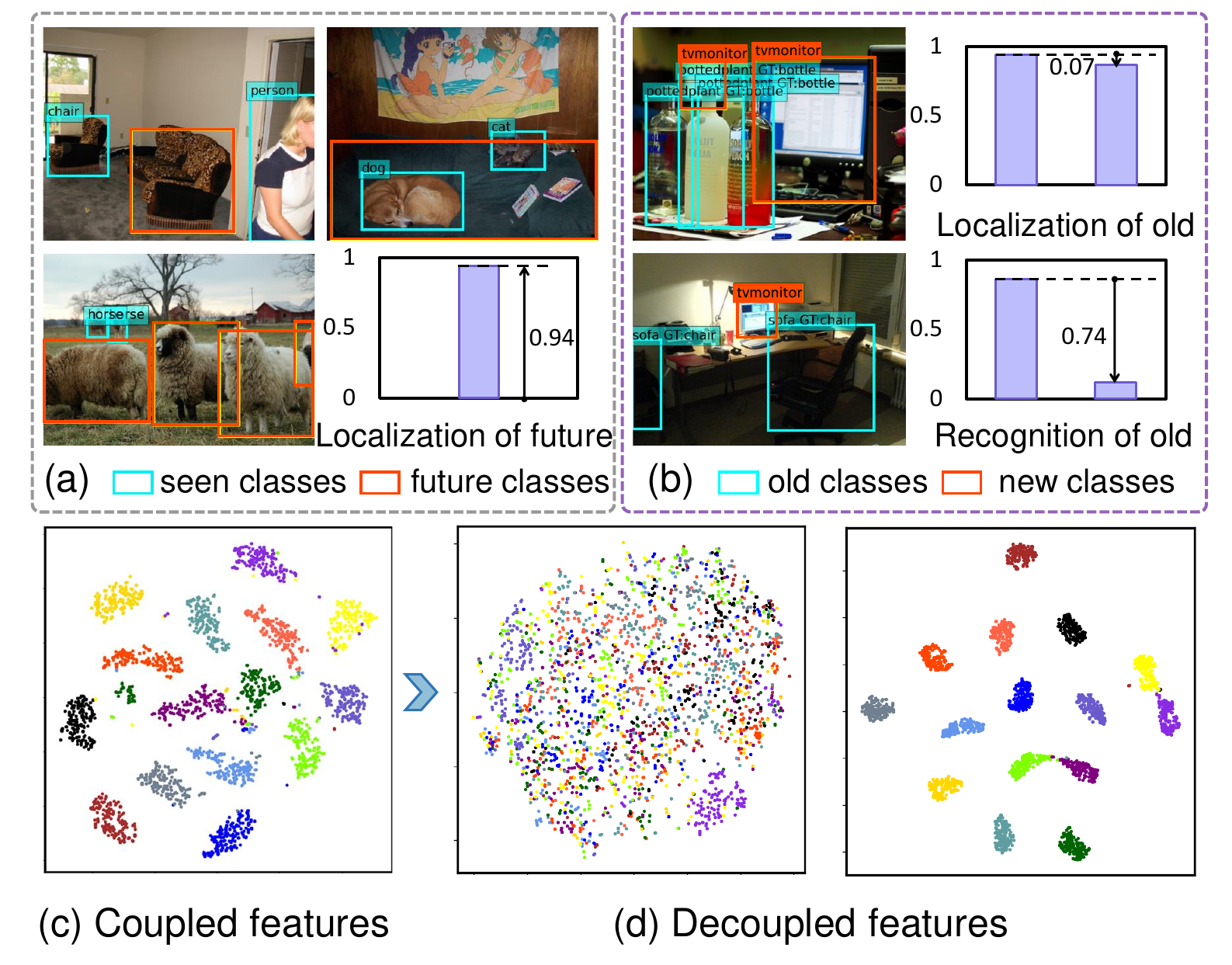}
\caption{{Forgetting imbalance within DETR-based IOD}. (a) Localization is class-agnostic and can generalize predicted boxes on future classes with a high recall of 94\%. (b) After fine-tuning on new data, localization is less-forgetting with recall of old classes slightly dropping from 94\% to 87\%, while average accuracy of recognition drops from 86\% to 12\%. (c) In original DETR, localization features are clearly influenced by recognition features and become category-specific. (d) After decoupling features, localization is obviously class-agnostic and we can focus on solving catastrophic forgetting on recognition.}
\label{cause-decouple}
\end{figure}
IOD suffers from catastrophic forgetting~\cite{kirkpatrick2017overcoming}, where models tend to erase prior knowledge when fine-tuning on new data. Due to privacy and security concerns, limited access to historical data makes training from scratch difficult. To address this, most methods preserve old knowledge through exemplar replay (ER)~\cite{liu2020multi, joseph2021towards} and knowledge distillation (KD)~\cite{zhou2020lifelong, yang2022rd}, which adopt CNN-based detectors containing numerous hand-crafted components as the basic framework~\cite{ren2015faster}.

Within transformer-based incremental detection (DETR) frameworks, established strategies like KD and ER are popularly used. For example, 
ACF~\cite{Kang_2023_ICCV} distills distance matrix and interactive features to attain class consistency. This inertial reuse has limited DETR-based IOD techniques to their full potential. \textit{Could we identify the underlying causes of catastrophic forgetting and suit the remedy to the case?} To answer this, we analyze \textit{forgetting} of classification and localization in DETR-based IOD, two primary tasks in object detection. We find that localization in DETR-based IOD is class-agnostic and less-forgetting while recognition suffers from severe forgetting, which we call the \textit{forgetting imbalance} between localization and recognition of DETR-based IOD. As shown in Figure~\ref{cause-decouple}(a), although the model is trained exclusively on seen categories (\textit{chair, person, horse, dog, cat}), it is capable of generating accurate localization for future categories (\textit{sofa, sheep, pottedplant}) with a high recall of 94\%. After finetuning on new data (Figure~\ref{cause-decouple}(b)), localization still provides accurate boxes for old categories while recognition has serious forgetting, such as \textit{bottle} being mistakenly identified as \textit{pottedplant}. The \textit{localization} recall of old classes drops slightly from 94\% to 87\% while average accuracy of \textit{recognition} drop from 86\% to 12\%. 

\begin{figure}[t]
\centering
\includegraphics[width=0.99\columnwidth]{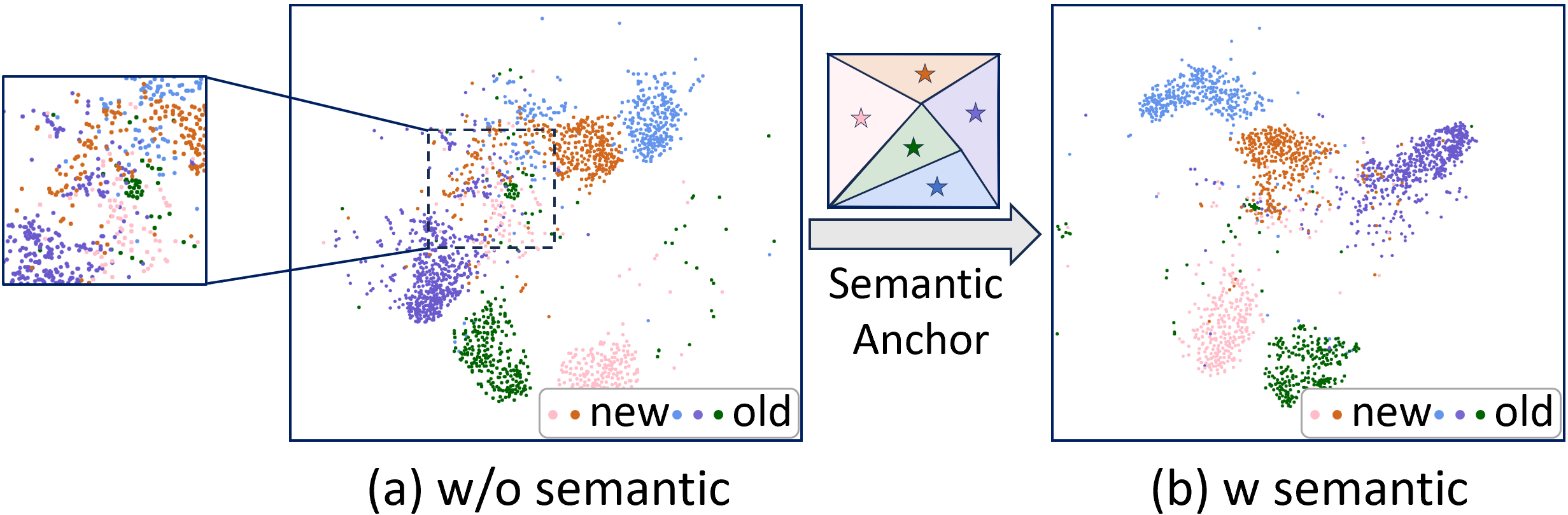}
\caption{(a) Feature ambiguity arises between new and old tasks due to feature drift. (b) With semantic that guides the optimization direction of each class, clear boundaries are effectively maintained.}
\label{fig:semantic}
\end{figure}

To address \textit{forgetting imbalance}, we break away from the mutual entanglement of forgetting and provide a reliable method for DETR-based IOD. We propose a Divide-and-Conquer Amnesia (DCA) strategy which decouples incremental detection into less-forgetting localization and fragile recognition. DCA first decodes class-agnostic location information. The recognition decoding process further determines the specific category of an object based on the predicted position. Figure~\ref{cause-decouple}(c\&d) compare localization features and recognition features in original DETR and after decoupling features, which illustrates that our ``dividing'' strategy frees the class-agnostic localization features from coupled features, and we can focus on solving recognition forgetting.

In the incremental recognition process, due to limited data access to past and future tasks, the semantic objectives for each task are optimized independently, focusing solely on the current task. The features of old classes experience significant drift or are completely overwritten, leading to feature ambiguity between new and old tasks, as shown in Figure~\ref{fig:semantic}(a), which impedes the model's compatibility with past, present, and future learning. Class semantic features from pre-trained language models (PLMs) anchor class representations, which constitute the semantic space, providing a unified optimization direction across tasks. 
Moreover, semantic understanding enables the model to comprehend relationships between new and old classes, promoting knowledge transfer from old to new tasks. We develop two mechanisms to embed semantic information. The first is duplex classifier fusion, which newly introduces a semantic classification head to calculate the similarities between recognition features and derived semantics.The second is {query-role embedding}, where semantic features of known tasks are incorporated into the recognition decoding process in the form of queries, implicitly introducing inter-class relationships through the attention mechanism.

To sum up, our contributions are as follows:
\begin{itemize}
    \item  We discover \textit{forgetting imbalance} in DETR-based IOD and further propose a Divide-and-Conquer Amnesia (DCA) strategy, which decouples incremental object detection into less-forgetting localization and class recognition to mitigate mutual interference.
    \item {To conquer severe forgetting in recognition, we effectively embed semantic knowledge from pre-trained language models to promote unified optimization across tasks by designing duplex classifier fusion and integrating semantic features into the recognition decoder in the form of queries, reducing feature shift in recognition.}
    \item Extensive experiments on two datasets demonstrate our proposed DCA achieves state-of-the-art performance compared to other exemplar-free methods and is particularly suitable for long-term incremental scenarios.
\end{itemize}
\section{Related Works}
\label{sec:formatting}
\noindent \textbf{Incremental Learning.}
Various approaches to mitigate catastrophic forgetting can be broadly divided into three categories: $i)$ Rehearsal-based methods~\cite{rebuffi2017icarl, huang2024kfc, zhu2025reshaping} involve storing a limited subset of old exemplars in memory buffers or utilizing a supplementary generator to synthesize pseudo samples for old data, which are then incorporated into training along with new data. $ii)$ Regularization-based methods~\cite{ hou2019learning, douillard2020podnet, huang2024etag} endeavor to design a loss function that penalizes changes in pivotal parameters during learning new tasks or utilize knowledge distillation to retain effective information acquired by previous models, such as output logits and intermediate features. $iii)$ Architecture-based methods~\cite{yan2021dynamically, wang2022learning, douillard2022dytox} modify the network architecture by adding sub-networks or experts when new tasks arrive while maintaining the previous network frozen.

\textbf{Incremental Object Detection.}
Previous methods focus on distilling knowledge such as outputs~\cite{shmelkov2017incremental}, feature maps~\cite{hao2019end} and various relations~\cite{peng2021sid, yang2022multi}, or replaying exemplars and intermediate features~\cite{acharya2020rodeo, joseph2021incremental, yang2023one, joseph2021towards, yang2023pseudo} to mitigate forgetting.
Moreover, IncDet~\cite{liu2020incdet} is based on parameter isolation, which mines and freezes important parameters. Built upon the DETR-based detection framework, CL-DETR distills knowledge from the most informative predictions of the old model and preserves the label distribution of the training set. ACF distills the distance matrix to preserve the inter-class discrimination and interactive features to maintain intra-class consistency. In this work, we uncover the causes of catastrophic forgetting in DETR-based IOD and propose a divide-and-conquer strategy to focus on the forgetting of class recognition. Moreover, our method has the advantage of exemplar-free overhead.
\begin{figure*}[t]
	\centering
\includegraphics[width=0.89\textwidth]{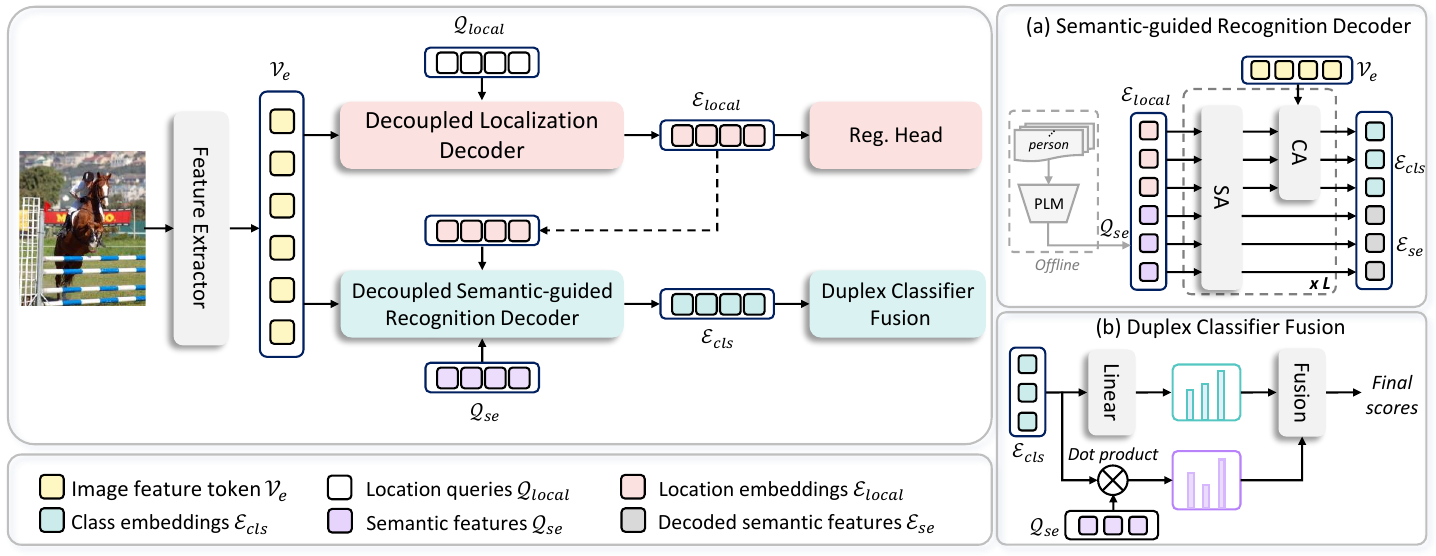}
    \caption{{Pipeline of DCA.} The extracted feature sequences $\mathcal{V}_{e}$ are first fed to decoupled localization decoder for object location embeddings $\mathcal{E}_{local}$, which are then sent to decoupled Semantic-guided Recognition Decoder to probe features to get class embeddings $\mathcal{E}_{cls}$. To integrate inter-class relationships, we embed semantic features $\mathcal{Q}_{se}$ from PLMs into recognition decoder in the form of queries and perform self-attention (SA) with location embeddings. To promote unified optimization across tasks, Duplex Classifier Fusion adds a semantic head to calculate similarities between $\mathcal{E}_{cls}$ and $\mathcal{Q}_{se}$ which are combined with the standard linear head to generate final recognition scores.}
	\label{framework}
\end{figure*}

\textbf{Incremental Learning with Foundation Models.} 
Recent trends in continual learning involve integrating pre-trained vision transformers with parameter-efficient fine-tuning techniques to adapt the model to downstream tasks. These techniques include prompt tuning~\cite{jia2022visual}, adapters~\cite{chen2022adaptformer}, LoRA~\cite{hulora}, etc. The core is to build additional learnable parameters or modules to guide the pre-trained representation and select appropriate prompts during reference. Another work leverages pre-trained vision-language models~\cite{radford2021learning} as foundation models~\cite{wang2023attriclip, smith2023coda, yang2024clip}, which can learn complex patterns between images and language. These methods learn lightweight adapter modules for both textual and visual pathways~\cite{gao2024clip} or soft prompts that serve as inputs to frozen visual and text encoders~\cite{zhou2022conditional} to capture task-specific information. However, directly integrating foundation models with DETR-based IOD incurs a lot of storage and computational overhead. In this work, we explore the rich inherent semantics mediated by pre-trained language models (PLMs) (either unimodal~\cite{kenton2019bert} or multimodal~\cite{radford2021learning}) to alleviate forgetting, which flexibly bridges IOD and foundation models.

\section{Preliminary}
\noindent \textbf{Problem Definition.}
The objective of IOD is to develop a unified detector that can adapt to newly encountered classes. Formally, given the dataset $\mathcal{D}\!=\!\left \{ (x,y) \right \} $, where $x$ is the image with corresponding object annotations $y$, the total category set of objects is $\mathcal{C}$. We partition the category set $\mathcal{C}$ into $T$ subsets, with $\mathcal{C} =\mathcal{C} _{1}\cup \cdots  \cup \mathcal{C} _{T}$, and corresponding sub-datasets constitute the training set for each phase. All label sets are mutually exclusive. At phase $t$, a group of classes $\mathcal{C}_{t}$ are exposed to the detector. For an image in the current phase, it may contain multiple objects from new classes $\mathcal{C}_{t}$ and other classes (old $\mathcal{C}_{1:t-1}$ and future $\mathcal{C}_{t+1:T}$). Only annotations belonging to $\mathcal{C}_{t}$ are preserved. After training phase $t$, the model is evaluated on all seen classes $\mathcal{C}_{1:t} = \mathcal{C} _{1}\cup \cdots \cup \mathcal{C} _{t}$. 

\textbf{Transformer-based Detectors.}
Following CL-DETR, we adopt Deformable DETR (short for D-DETR) as the baseline architecture, which consists of a CNN backbone, a transformer encoder-decoder, and predictors of class and bounding box. The CNN backbone and transformer encoder aim to extract enhanced feature sequences $\mathcal{V}_{\mathrm{e}}$. The decoder takes features and a set of learnable object queries $\mathcal{Q}=\left \{ q_{i}\in \mathds{R}^{d}|i=1,...,N \right \} $ as input and outputs the object embeddings $\mathcal{E}$, followed by predictors to parse out bounding boxes $\mathcal{B}$ and classes $\mathcal{P}$. The process can be expressed as:
\begin{align}
    \mathcal{E}&=\mathrm{Decoder}(\mathcal{V}_{e},\mathcal{Q}), \\
    \mathcal{B}=&\mathrm{Reg}(\mathcal{E}), \mathcal{P}=\mathrm{Cls}(\mathcal{E}).
\label{eq:detr}
\end{align}
Hungarian algorithm is used to find a bipartite matching between ground truths and predictions of object queries.

\section{Methodology}
\label{sec:methods}
\subsection{Overview}
\noindent Considering the above observations, we redesign the decoding process of DETR-based IOD into a localization-then-recognition process to maintain the less-forgetting localization well and focus on conquering fragile recognition. Figure~\ref{framework} shows the overall framework of DCA, which retains the backbone and transformer encoder of D-DETR for extracting feature sequences $\mathcal{V}_{e}$ while modifying the transformer decoder. First, randomly initialized learnable location queries $\mathcal{Q}_{local}$ along with feature sequences $\mathcal{V}_{e}$ are sent to \textit{Decoupled Localization Decoder} to obtain location embeddings $\mathcal{E}_{local}$, followed by a regression head for predicting wide-coverage boxes. Next, for prior locations and one-to-one bipartite matching between localization and recognition decoding, class queries $\mathcal{Q}_{cls}$ are initialized with location embeddings. To integrate inter-class relationships, semantic features $\mathcal{Q}_{se}$, obtained by sending class names into PLMs, enter decoder in the form of queries. $\mathcal{Q}_{cls}$, concatenated with $\mathcal{Q}_{se}$, are updated through \textit{Decoupled Semantic-guided Recognition Decoder} to get class embeddings $\mathcal{E}_{cls}$. \textit{Duplex Classifier Fusion} predicts final recognition scores, which adds the semantic-based head to build a unified feature optimization space within and across tasks.
{\subsection{Decoupled Localization and Recognition}}
\label{disentangled} 
\noindent {As illustrated in Eq.(1)(2), the original coupled architecture of DETR shares localization and recognition features $\mathcal{E}$ which are sent to Reg. head and Cls. head for predicted boxes and scores.  
As observed in Figure~\ref{cause-decouple}(a)(b), since localization is less-forgetting and can also generalize to future classes similar to seen classes, directly put constraints on these coupled features to preserve old knowledge, such as knowledge distillation, will affect the generalization of localization, thereby reducing plasticity. To avoid mutual interference, we split the original transformer decoder into decoupled localization decoder and recognition decoder.}

{As shown in Figure~\ref{framework}, the feature extractor retained in D-DETR extract enhanced image feature sequences $\mathcal{V}_{e}$. For localization, N randomly initialized location queries $\mathcal{Q}_{local}\in \mathbb{R}^{N\times d}$ and visual features $\mathcal{V}_{e}$ are input to decoupled localization decoder to obtain location embeddings $\mathcal{E}_{local}\in \mathbb{R}^{N\times d}$ of foreground objects, followed by Reg. head for box coordinates $\mathcal{B} =\left \{ \bar{b}_{1},\dots ,\bar{b}_{N} \right \} \in \mathbb{R}^{N\times 4}$. The localization decoder is composed of $L$ blocks of transformer layers, following D-DETR. For recognition, it aims to classify the objects in predicted boxes. We use all location embeddings and image feature tokens as input for decoupled recognition decoder. To integrate inter-class relationships, semantic features $\mathcal{Q}_{se}$ from pre-trained language models are also fed forward to class recognition decoder (see the next section for details). For each location embedding, we get its class embedding $\mathcal{E}_{cls}\in \mathbb{R}^{N\times d}$. Then, duplex classifier fusion head is applied to $\mathcal{E}_{cls}$ to get class probabilities $\mathcal{P}=\left \{ \bar{p}_{1},\dots ,\bar{p}_{N} \right \} \in \mathbb{R}^{N\times K}$ after $\mathrm{sigmoid(\cdot)}$, where $K=|\mathcal{C}_{1:t}|$ is the number of all seen classes. Compared with original coupled decoding in Eq.(1)(2), our decoupled decoding can be expressed as:}
\begin{align}
    \mathcal{E} _{local} &= \mathrm{{Decoder}\_{Local}} (\mathcal{V}_{\mathrm{e}},\mathcal{Q}_{{local}} ),\\
        \mathcal{Q} _{{ cat}} &= \mathrm{Concat}(\mathcal{E}_{{local}},\mathcal{Q}_{{se}}),\\
\mathcal{E}_{{cls}}&= \mathrm{{Decoder\_{Cls}}}(\mathcal{V}_{\mathrm{e}},\mathcal{Q} _{{cat}}),\\
    \mathcal{B}=\mathrm{R}&\mathrm{eg}(\mathcal{E}_{local}), \mathcal{P}=\mathrm{DCF}(\mathcal{E}_{cls}),
\end{align}
where $\mathrm{Decoder\_Local}$, $\mathrm{Decoder\_Cls}$ denote the localization decoder and recognition decoder respectively, which share parameters to have a comparable parameters count to other baselines. $\mathrm{DCF}$ means duplex classifier fusion. After obtaining the optimal assignment $\bar{\sigma}$ between prediction results and ground-truth annotations ${y} \!=\! \left \{ \left ( {c}_{i},{b_{i}}\right )\right \}_{i=1}^{N} $, we optimize the model by minimizing the following detection loss:
\begin{align}
    \mathcal{L}_{{det}}\!=\!\sum_{i=1}^{{N}}\!\left [ -\mathrm{log} \bar{p}_{\bar{\sigma}(i) } (c_{i}) \!+\! \mathds{1}_{\left \{ c_{i}\ne \phi \right \}}\mathcal{L}_{{box} }( b_{i},\!\bar{b}_{\bar{\sigma }(i)}\right ],
\end{align}
where $\mathcal{L}_{{box}}\!=\! \lambda _{{iou} }\mathcal{L} _{{iou} }(b_{i}, \bar{b}_{\bar{\sigma} (i) } )\! +\! \lambda_{\mathrm{\mathcal{L} 1} }\!\left \| b_{i} \!- \!\bar{b}_{\bar{\sigma} (i) } \right \|_{1}$ and $\mathds{1}(\cdot )$ is the indicator function. Comparison in Figure~\ref{cause-decouple}(c)(d) illustrates that our decoupling operation isolates class-agnostic localization features and recognition features.

\subsection{Class Recognition with Semantic Guidance}
\noindent {To conquer recognition forgetting, we explore the semantic knowledge from pre-trained language models (PLMs) to guide class recognition, which helps provide a unified optimization target for the entire incremental process. We first integrate semantic features into the recognition decoder in the form of queries to consider high-level semantic relationships between objects, forming decoupled semantic-guided recognition decoder. Then, we leverage the well-structured feature space spanned by these semantic features to design the duplex classifier fusion, reducing feature drift.}

\textbf{Semantic-guided Recognition Decoder.}
\label{semantic}
It consists of $L$ attention blocks, each comprising a self-attention (SA) and a cross-attention (CA) layer. In the self-attention module, each query (here also referred to as location embeddings from the localization decoder) updates its representation by aggregating information from all other queries, which allows the model to reason about the relationships between different queries. It helps the decoder refine the predictions by considering the interaction among all predicted objects, such as reducing duplicate detections and adjusting bounding boxes based on the presence of other objects. We concatenate the semantic features with queries, and these enriched queries serve as the input to the decoder’s self-attention layer. By integrating semantic features, each query becomes more robust in its representation, which is not limited to just physical attributes of objects (like spatial and appearance-based information) but also high-level semantic relationships between objects. It is beneficial to IOD for two reasons. Firstly, semantic features allow the model to generalize better to new classes. The model leverages shared semantic traits across different but related classes, facilitating quicker and more accurate adaptation to new data. Secondly, semantic queries ensure that the model maintains a consistent understanding and representation of old classes over multiple incremental steps by injecting the semantic features into class queries, which effectively mitigates the semantic drift of old classes.

As shown in Figure~\ref{framework}(a), semantic features $\mathcal{Q}_{se}$ are generated by filling known class names $\mathcal{C}_{1:t}$ into a template and feeding to the pre-trained language model. Class queries ($i.e.$, $\mathcal{E}_{local}$) and semantic features are input to self-attention to deduce relations among objects and simultaneously infuse object features with prior semantic information. The enriched queries interact with image features in the cross-attention. The $l$-th semantic-guided decoder block can be illustrated as follows:
\begin{align}
    (\mathcal{E} _{{cls}}^{l'}, \mathcal{E} _{{se}}^{l})&=\mathrm{SA} (\mathcal{E} _{{cls}}^{l-1 }, \mathcal{E} _{{se}}^{l-1 }),\\
    \mathcal{E} _{{cls}}^{l}&=\mathrm{CA} (\mathcal{E} _{{cls}}^{l' }, \mathcal{V}_{{e}}),
\end{align}
where $\mathcal{E}^{l'}_{cls}$ is the intermediate output by SA, and $\mathcal{E}^{l}_{cls}$ and $\mathcal{E}^{l}_{se}$ are decoded class embeddings and semantic features of $l$-th block. To prevent semantic drift when decoded semantic features are fed into the next transformer block, we impose the semantic consistency loss to keep consistent with the original semantic features $\mathcal{Q}_{se}$ by cosine similarities $\mathrm{cos(\cdot)}$ across all semantic-guided recognition blocks:
\begin{equation}
\begin{split}
    \mathcal{L}_{{cons} } = \sum_{l=1 }^{L} \left (1-\mathrm{cos}(\mathcal{Q}_{{se}}, \mathcal{E}^{l} _{{se}})  \right ). 
\end{split}
\end{equation}

\textbf{Duplex Classifier Fusion.}
\label{duplex}
Previous optimization of classifiers focuses narrowly on the current training data. Due to limited access to old data and the unpredictability of new data, new weights of the model overwrite previous knowledge, resulting in distortion and overlap of feature spaces between old and new tasks. It is difficult for the model to be compatible with both old and new classes.
Semantic-based classifier offers a more robust solution, which is endowed with a well-constructed global feature space, spanned by semantic features derived from all known class names. Due to being pre-trained on a large dataset, this model gains a conceptual understanding of data and can learn universal semantic relationships both within and across tasks. When new classes arrive, pre-defined new semantic features can be registered without compromising the integrity of the old feature space, thereby mitigating catastrophic forgetting. Integrating the strengths of both, we propose duplex classifier fusion for the balance between plasticity and stability.

As illustrated in Figure~\ref{framework}(b), class embeddings $\mathcal{E}_{cls}$ are fed through the linear layer to get class probabilities $\mathcal{H}=\left \{ \bar{h}_{1},\dots ,\bar{h}_{N} \right \} \in \mathbb{R}^{N\times K}$ after $\mathrm{sigmoid(\cdot)}$ activation. On the other hand, a projection layer is used to map class embeddings to semantic space for projected class embeddings $\mathcal{E}_{proj}$ and then semantic class probabilities $\mathcal{S}=\left \{ \bar{s}_{1},\dots ,\bar{s}_{N} \right \} \in \mathbb{R}^{N\times K}$ are obtained by calculating similarities between projected class embeddings and semantic features of all known classes. Here, we directly employ a weighted approach (weight $\beta=0.5$) to fuse these probabilities to get the overall classification probabilities $\mathcal{P}=\left \{ \bar{p}_{1},\dots ,\bar{p}_{N} \right \} \in \mathbb{R}^{N\times K}$ for training and inference:
\begin{align}
    \bar{p}_{i}=\beta \cdot \bar{h}_{i} + (1-\beta)\cdot \bar{s}_{i}
\end{align}

\subsection{Hybrid Knowledge Distillation}
\label{hybrid}
\noindent Given that the current training data contains unlabeled old objects, we add pseudo labels to foreground predictions of the old model as supplement supervision, solving background interference. However, the shortage of old data leads to the bias towards new ones. We employ hybrid knowledge distillation to retain a diverse range of information from various perspectives and levels. Firstly, we constrain the detection outputs including class probabilities and boxes:
\begin{align}
\label{Eq:logkd}
        \mathcal{L}^{{kd} } _{{out}} \!= \!\mathcal{L}_{{mse} }(\bar{p}^{{new} } ,\bar{p}^{{old} } )\! +\!\mathcal{L}_{box }(\bar{b}^{{new} },\bar{b}^{{old} }),
\end{align}
where $(\bar{p}^{{new} },\bar{b}^{{new} })$, $(\bar{p}^{{old} },\bar{b}^{{old} })$ represent prediction results of new and old model respectively. Meanwhile, significant drift in image features can adversely affect the recognition decoding, leading to a marked reduction in class distinction. Class embeddings are also susceptible to severe deviation that exacerbate forgetting. To counteract this, we preserve visual features learned by the old model, encompassing both the encoded features $\mathcal{V}_{e}$ and decoded class embeddings $\mathcal{E}_{cls}$. To improve the plasticity for new classes, we use pseudo-labels of old classes as instance-level masks for selection distillation:
\begin{align}
    \mathcal{L}^{{kd} } _{{vis}}&=\mathcal{G}   (\mathcal{V}_{{e}})+\mathcal{G}   (\mathcal{E}_{{cls}}),
\end{align}
where $\mathcal{G}(f) \!=\!\frac{1}{N^{{old} }}\sum_{j=1}^{N^{{old} } } A_{ij}\left \| f^{{new} }_{ij}\!-\!f^{{old} }_{ij} \right \|_{1}$, and $N^{{old}}$ is the sum of old pseudo-boxes. To prevent the projected features $\mathcal{E}_{{proj}}$ shifting, we impose distillation:
\begin{equation}
    \label{Eq:se-featkd}
    \begin{aligned}
    \mathcal{L}^{{kd} } _{{proj}}=\mathcal{G}   (\mathcal{E}_{{proj}}).
    \end{aligned}
\end{equation}
Then the hybrid knowledge distillation is $\mathcal{L}_{{hkd} }= \mathcal{L} ^{{kd} }_{{out} }+\mathcal{L} ^{{kd} }_{{vis} }+\mathcal{L} ^{{kd} }_{{proj} }$. To sum up, the overall loss for training the new model is given by ($\mathcal{L}_{hkd}=0$ when training base model):
\begin{align}
        \mathcal{L} _{{all} } &= \mathcal{L} _{{det} } + \mathcal{L}_{cons}+\mathcal{L}_{{hkd} }.
\end{align}

\begin{table*}[t]
\small
\setlength{\tabcolsep}{0.5mm}
	\begin{center}
			\begin{tabular}{clccccccccccccccccccccc}
				\toprule
				\makebox[0.1mm][c]{\multirow{9}{*}{\rotatebox{90}{10+10 Setting}}} &{Method} & aero & bike & bird & boat & bottle & bus & car & cat & chair & {cow} & table & dog & horse & mbike & person & plant & sheep & sofa & train & {tv} &\multicolumn{1}{c}{ mAP}\\
				\cmidrule{1-23}
     & Faster ILOD&
                72.8 &75.7 &71.2&60.5 &61.7&70.4& 83.3& 76.6& 53.1 &{72.3} &\cellcolor{gray!12}36.7 &\cellcolor{gray!12}70.9& \cellcolor{gray!12}66.8& \cellcolor{gray!12}67.6& \cellcolor{gray!12}66.1&\cellcolor{gray!12}24.7 &\cellcolor{gray!12}63.1&\cellcolor{gray!12}48.1 &\cellcolor{gray!12}57.1 &\cellcolor{gray!12}{43.6}&\multicolumn{1}{c}{62.1}\\
    &MMA&\multicolumn{10}{c}{69.3}& \multicolumn{10}{c}{\cellcolor{gray!12}63.9}& \multicolumn{1}{c}{66.6}\\
    &ORE-EBUI*&
                63.5&70.9&58.9&42.9&34.1 &76.2& 80.7&76.3& 34.1& {66.1} &\cellcolor{gray!12}56.1&\cellcolor{gray!12}70.4 &\cellcolor{gray!12}80.2&\cellcolor{gray!12}72.3&\cellcolor{gray!12}81.8& \cellcolor{gray!12}42.7 &\cellcolor{gray!12}71.6 &\cellcolor{gray!12}68.1 &\cellcolor{gray!12}77 &\cellcolor{gray!12}{67.7}&\multicolumn{1}{c}{64.5}\\
    &OW-DETR*&
                61.8 &69.1 &67.8 &45.8 &47.3 &78.3 &78.4 &78.6 &36.2& {71.5} &\cellcolor{gray!12}57.5 &\cellcolor{gray!12}75.3 &\cellcolor{gray!12}76.2 &\cellcolor{gray!12}77.4 &\cellcolor{gray!12}79.5 &\cellcolor{gray!12}40.1& \cellcolor{gray!12}66.8& \cellcolor{gray!12}66.3& \cellcolor{gray!12}75.6& \cellcolor{gray!12}{64.1} &\multicolumn{1}{c}{65.7}\\
    &CL-DETR\dag &55.0 & 63.0 & 51.1 & 35.3 & 36.2 & 48.2 & 60.5 & 45.9 & 29.8 & 21.1 & \cellcolor{gray!12}61.6 & \cellcolor{gray!12}72.1 & \cellcolor{gray!12}75.0 & \cellcolor{gray!12}75.4 & \cellcolor{gray!12}75.1 & \cellcolor{gray!12}38.5 & \cellcolor{gray!12}65.4 & \cellcolor{gray!12}62.2 & \cellcolor{gray!12}76.4 &\cellcolor{gray!12} 68.4 & 55.8\\
    &PROB\dag &70.4 &75.4& 67.3 &48.1& 55.9 &73.5 &78.5& 75.4 &42.8& 72.2 &\cellcolor{gray!12}64.2& \cellcolor{gray!12}73.8 &\cellcolor{gray!12}76.0 &\cellcolor{gray!12}74.8 &\cellcolor{gray!12}75.3 &\cellcolor{gray!12}40.2& \cellcolor{gray!12}66.2 &\cellcolor{gray!12}73.3 &\cellcolor{gray!12}64.4 &\cellcolor{gray!12}64.0& 66.5\\
    &ACF&\multicolumn{10}{c}{67.0}& \multicolumn{10}{c}{\cellcolor{gray!12}70.1}& \multicolumn{1}{c}{\underline{68.6}}\\
		      \cmidrule{2-23}
        
&\textbf{DCA} &	79.7 &82.3 &71.0 & 61.3 & 65.1 & 80.0 & 86.9 &77.3& 56.2&{73.1}&\cellcolor{gray!12}65.8&\cellcolor{gray!12}81.8&\cellcolor{gray!12}85.2&\cellcolor{gray!12}81.5 & \cellcolor{gray!12}80.6 & \cellcolor{gray!12}45.6 &\cellcolor{gray!12}70.0 & \cellcolor{gray!12}68.8 & \cellcolor{gray!12}82.7 & \cellcolor{gray!12}{73.0} & {\textbf{73.4}}\\
				\midrule
				\midrule
    \makebox[0.1mm][c]{\multirow{9}{*}{\rotatebox{90}{15+5 Setting}}} & Faster ILOD&
				66.5 &78.1&71.8 &54.6& 61.4 &68.4 &82.6 &82.7&52.1& 74.3& 63.1 &78.6& 80.5 &78.4& {80.4} &\cellcolor{gray!12}36.7 &\cellcolor{gray!12}61.7&\cellcolor{gray!12}59.3 &\cellcolor{gray!12}67.9& \cellcolor{gray!12}{59.1} &\multicolumn{1}{c}{67.9}\\
    &MMA&\multicolumn{15}{c}{73.0}& \multicolumn{5}{c}{\cellcolor{gray!12}60.5}& \multicolumn{1}{c}{69.9}\\
    &ORE-EBUI* & 75.4& 81& 67.1 &51.9 &55.7 &77.2 &85.6 &81.7& 46.1& 76.2 &55.4& 76.7& 86.2 &78.5 &{82.1}& \cellcolor{gray!12}32.8& \cellcolor{gray!12}63.6 &\cellcolor{gray!12}54.7 &\cellcolor{gray!12}77.7&\cellcolor{gray!12}{64.6} &\multicolumn{1}{c}{68.5}\\
    &OW-DETR*  &
                77.1& 76.5 &69.2 &51.3 &61.3& 79.8& 84.2& 81.0& 49.7 &79.6& 58.1 &79.0& 83.1 &67.8& {85.4}& \cellcolor{gray!12}33.2 &\cellcolor{gray!12}65.1 &\cellcolor{gray!12}62.0& \cellcolor{gray!12}73.9&\cellcolor{gray!12}{65.0} &\multicolumn{1}{c}{69.1}\\
    &CL-DETR\dag  & 64.0 & 60.0 & 56.6 & 38.5 & 44.9 & 42.6 & 68.7 & 55.0 & 36.4 & 50.9 & 37.8 & 59.2 & 76.0 & 71.4 & 72.2 & \cellcolor{gray!12}32.1 & \cellcolor{gray!12}32.1 & \cellcolor{gray!12}34.6 & \cellcolor{gray!12}47.7 & \cellcolor{gray!12}52.3 & 51.6\\
    &PROB* & 77.9& 77.0& 77.5 &56.7& 63.9 &75.0 &85.5& 82.3& 50.0& 78.5 &63.1 &75.8 &80.0& 78.3 &77.2& \cellcolor{gray!12}38.4 &\cellcolor{gray!12}69.8 &\cellcolor{gray!12}57.1&\cellcolor{gray!12}73.7& \cellcolor{gray!12}64.9 &70.1 \\
    &ACF&\multicolumn{15}{c}{71.6}& \multicolumn{5}{c}{\cellcolor{gray!12}65.9}& \multicolumn{1}{c}{\underline{70.2}}\\
                \cmidrule{2-23}
&\textbf{DCA}&75.0 & 84.1 & 78.2 & 64.8 & 63.8 & 69.9 & 87.6 & 86.1 & 58.6 & 72.2 & 73.0 & 83.7 & 83.4 & 82.6 & 83.8 & \cellcolor{gray!12}36.7 & \cellcolor{gray!12}52.3 & \cellcolor{gray!12}53.0 & \cellcolor{gray!12}72.6 & \cellcolor{gray!12}61.7 & \textbf{71.2} \\
            \midrule
			\midrule
    \makebox[6mm][c]{\multirow{9}{*}{\rotatebox{90}{19+1 Setting}}} & Faster ILOD&
                64.2& 74.7 &73.2& 55.5& 53.7 &70.8 &82.9 &82.6 &51.6 &79.7& 58.7 &78.8 &81.8& 75.3 &77.4 &43.1& 73.8 &61.7 &{69.8}& \cellcolor{gray!12}{61.1}& {68.5}\\
    &MMA&\multicolumn{19}{c}{71.1}& \cellcolor{gray!12}{63.4}&\multicolumn{1}{c}{{70.7}}\\
    &ORE-EBUI*  &
                67.3 &76.8 &60 &48.4& 58.8& 81.1& 86.5 &75.8& 41.5& 79.6& 54.6 &72.8&85.9 &81.7& 82.4 &44.8& 75.8& 68.2&  {75.7}& \cellcolor{gray!12}{60.1}&{68.8}\\
    &OW-DETR*  &
                70.5& 77.2 &73.8 &54.0 &55.6 &79.0 &80.8& 80.6& 43.2 &80.4& 53.5& 77.5& 89.5 &82.0& 74.7& 43.3& 71.9& 66.6& {79.4}& \cellcolor{gray!12}{62.0} &{69.8}\\
    &CL-DETR\dag  &63.1 & 61.0 & 56.3 & 42.2 & 50.9 & 48.5 & 72.5 & 61.3 & 44.3 & 68.2 & 47.3 & 67.3 & 66.4 & 63.9 & 71.9 & 37.5 & 66.9 & 44.6 & 62.2 & \cellcolor{gray!12}45.7 & 57.1 \\
    &PROB* & 80.3 &78.9& 77.6 &59.7& 63.7 &75.2 &86.0 &83.9& 53.7 &82.8& 66.5& 82.7 &80.6 &83.8 &77.9 &48.9& 74.5 &69.9& 77.6& \cellcolor{gray!12}48.5 &\underline{72.6} \\
    &ACF&\multicolumn{19}{c}{71.9}& \cellcolor{gray!12}{66.9}&\multicolumn{1}{c}{70.6}\\
                \cmidrule{2-23}
&\textbf{DCA}&73.9&83.2&79.6&60.1&63.6&75.0&86.7&85.6&61.2&79.8&71.3&83.2&85.3&82.4&83.4&49.9&76.3&70.0&79.2&\cellcolor{gray!12}61.1&{\textbf{74.5}}\\
				\bottomrule
\end{tabular}
\end{center}
\caption{Per-class average precision on VOC test dataset where $10$, $5$ or $1$ classes are added at once. Best among rows \textbf{in bold} and second best are \underline{underlined}. Methods with * store old-class data or use extra wild data and with † come from re-implementation.}
\label{table:voc-two-step}
\end{table*}

\section{Experiments}
\subsection{Experimental Setups}
\noindent \textbf{Datasets and Metrics.}
We evaluate on two widely used datasets: PASCAL VOC~\cite{everingham2007pascal} and MS COCO~\cite{lin2014microsoft}. VOC contains 20 foreground classes, while COCO covers 80 object categories. $AP_{50}$ and $AP$ are reported for metrics. To measure the gap between an incremental model response and the ideal setting, we use the absolute gap (AbsGap) and relative gap (RelGap).

\textbf{Incremental Protocols.}
We simulate diverse learning scenarios. For VOC, we consider three different settings, where a group of classes (10, 5 and last class) are introduced incrementally to the detector. For COCO, we conduct experiments under 70+10, 60+20, 50+30 and 40+40 settings. To increase the task difficulty, multi-step settings are evaluated where base model is trained with 40 classes and 20 or 10 classes are added in each of the following phases.

\textbf{Implementation Details.}
The architecture of DCA detector is an adaptation of D-DETR, which leverages ResNet-50~\cite{he2016deep} pre-trained on ImageNet~\cite{deng2009imagenet} as the backbone. Following CL-DETR, we use the standard configurations without iterative bounding box refinement and the two-stage mechanism. For the shared decoder, the number of layers is set to $L\!=\!6$ and the number of location queries $N\!=\!100$. During inference, top-50 high-scoring detections per image are used for evaluation. 
In DCA, we use CLIP text encoder as the language model to generate semantic features in an offline manner. Our codes are available at https://github.com/InfLoop111/DCA.

\subsection{Comparison with State-Of-The-Art Methods}
\begin{table}[t]
\small
\setlength{\tabcolsep}{0.6mm}
	\begin{center}
\begin{tabular}{l|cc|cc|cc|cc}
    \toprule
\multirow{2}{*}{Method} & \multicolumn{2}{c|}{\textbf{70+10}} & \multicolumn{2}{c|}{\textbf{60+20}} & \multicolumn{2}{c|}{\textbf{50+30}} & \multicolumn{2}{c}{\textbf{40+40}} \\ \cmidrule{2-9} 
                        &                                      AP          & AP50         & AP          & AP50         & AP          & AP50         & AP         & AP50         \\ \midrule
LwF                        &  7.1 &12.4           &    5.8& 10.8          &    5.0& 9.5&  17.2& 25.4                      \                                     \\
RILOD                 &     24.5& 37.9        &    25.4 &38.8          &   28.5 &43.2 &29.9 &45.0\\
SID                  &     32.8 & 49.0        &    32.7 &49.8          &   33.8 &51.0          &        34.0 &51.4                      \\
ERD                &     34.9 &51.9        &  35.8 &\underline{52.9}            &  36.6 &\underline{54.0}           &    36.9 &54.5                                                \\ \cmidrule{1-9}
CL-DETR            &   35.8& 53.5          &    -&-          &   -&-          & 39.2& 56.1                                              \\
CL-DETR*             &   \underline{40.4}& \underline{58.0}          &  -&-            &    -&-         &   \underline{42.0}& \textbf{60.1}                                                             \\
ACF                &   37.6 & -          &     \underline{38.3} & -         &                \underline{38.8} & -                     &  39.8 & -                                      \\
\textbf{DCA}          &  \textbf{41.3}&\textbf{59.2}           &  \textbf{41.9 }&\textbf{ 54.8}                          &  \textbf{39.9 }&\textbf{ 56.1}           & \textbf{42.8}&\underline{58.4}                                       \\ \bottomrule
\end{tabular}
\end{center}
\caption{Results ($AP/AP_{50}$) on COCO two-step setting. - represents no corresponding results in the original paper.}
\label{table:coco-two-step}
\end{table}

\noindent \textbf{One-step incremental settings.}
As reported in Table~\ref{table:voc-two-step}, DCA outperforms all previous non-exemplar methods in different data split setting. In particular, in 10+10 setting, DCA significantly improves the best non-exemplar method ACF by 4.8\%. DCA even exceeds all exemplar-based methods which store samples. Take the result of 19+1 setting as an example, we surpass the second best method PROB by 1.9\%, which validates the superiority of DCA. This phenomenon remains the same in COCO dataset with more categories.
\begin{table}[t]
\small
    \setlength\tabcolsep{0.5mm}
	\begin{center}
			\begin{tabular}{l|c|cc|c|c}
    \toprule
        Method& \textbf{(1-40)}&\textbf{+(40-60)}&\textbf{+(60-80)}& \textbf{AbsGap$\downarrow$} & \textbf{RelGap$\downarrow$}\\
         \midrule
         CF        & 45.7/ 66.3  & 10.7/ 15.8 & 9.4/ 13.3 & 30.8/ 45.0  & 0.77/ 0.77 \\
         RILOD       & 45.7/ 66.3  & 27.8/ 42.8 & 15.8/ 4.0 & 24.4/ 54.3  & 0.61/ 0.93\\
         SID         & 45.7/ 66.3  & 34.0/ 51.8 & 23.8/ 36.5 & 16.4/ 21.8 & 0.41/ 0.37\\
         ERD  & 45.7/ 66.3  & 36.7/ 54.6 & 32.4/ 48.6 & 7.8/ 9.7   & 0.19/ 0.17 \\
         ACF & 48.0/ - & 39.3/ - & 36.6/ - & 3.7/ -& 0.09/ - \\
\textbf{DCA}                    & 48.0/ 68.9  & \textbf{42.7/ 59.6} & \textbf{40.3/ 54.1} & \textbf{2.3/ 7.3}  & \textbf{0.05/ 0.12}\\
         \bottomrule
    \end{tabular}
	\end{center}
 \caption{Results ($AP/AP_{50}$) under two-step setting on COCO where (a-b) is the base normal training for classes a-b and +(c-d) is the incremental training for classes c-d.}
		\label{table:coco-20-2}
\end{table}
\begin{table}[t]
\small
    \setlength\tabcolsep{0.05mm}
	\begin{center}
			\begin{tabular}{l|cccc|c|c}
    \toprule
Method&\textbf{+(40-50)}& \textbf{+(50-60)}& \textbf{+(60-70)}& \textbf{+(70-80)}& \textbf{AbsGap$\downarrow$} & \textbf{RelGap$\downarrow$} \\
         \midrule
         \multirow{2}{*}{CF}     & {5.8} &5.7    &6.3  & 3.3  &36.9   &0.92   \\
             & 8.5   & 8.3   &8.5  &  4.8 &53.5  &0.92   \\ \cmidrule{1-7}
         \multirow{2}{*}{RILOD}   &25.4 &11.2 &10.5&8.4 &31.8  &0.79  \\
         &38.9 &17.3 &15.6&12.5 &45.8  &0.79  \\ \cmidrule{1-7}
         \multirow{2}{*}{SID}   &34.6  &24.1 &14.6&12.6 &27.6 &0.69 \\
          &52.1  &38.0 & 23.0& 23.3 & 35.0 & 0.60 \\ \cmidrule{1-7}
         \multirow{2}{*}{ERD}  &36.4  &30.8 &26.2&20.7 &19.5 &0.49 \\
         &53.9  &46.7 & 39.9& 31.8 & 26.5 & 0.46 \\ \cmidrule{1-7}
         \multirow{2}{*}{ACF}  & 39.1 & 35.4&32.0 &30.3& 10.0 & 0.25\\
         & - & -&- &- & - & -\\ \cmidrule{1-7}
         
\multirow{2}{*}{\textbf{DCA}}             & \textbf{44.0}  &\textbf{41.1} &\textbf{39.2} &\textbf{37.2} &\textbf{5.4} &\textbf{0.13} \\
& \textbf{ 61.2}  &\textbf{ 56.5} &\textbf{ 53.8} &\textbf{49.6} &\textbf{ 11.8} &\textbf{0.19}\\
         \bottomrule
    \end{tabular}
	\end{center}
 \caption{Results ($AP/AP_{50}$) under COCO four-step setting.}
		\label{table:coco-10-4}
\end{table}
\begin{table}[t]
\small
\setlength\tabcolsep{0.8mm}
	\begin{center}
			\begin{tabular}{c|cccc|cccc}
				\toprule
Row & DLR & SRD & DCF & HKD & All (gain $\bigtriangleup $) & Old & New & Avg. \\ \midrule
1   &     &     &     &     &     62.5 {(+0.0)}&64.8&55.4&60.1      \\
2   &   \checkmark  &     &     &     &     65.7 {(+3.2)}&68.9&56.1&62.5      \\
3   &     &  \checkmark   &     &     &    64.1 {(+1.6)}&67.3&54.7&61.0      \\
4   &   \checkmark  &  \checkmark   &     &     &     67.9 {(+5.4)}&70.8&59.0&64.9    \\
5   &  \checkmark   &  \checkmark   &   \checkmark  &     &     68.6 {(+6.1)}&72.1&57.0&64.6     \\
\cellcolor{gray!12}6   &   \cellcolor{gray!12}\checkmark  &  \cellcolor{gray!12}\checkmark   &  \cellcolor{gray!12}\checkmark   &  \cellcolor{gray!12}\checkmark   &     \textbf{71.2} \cellcolor{gray!12} {(+8.7)}&\cellcolor{gray!12}{76.5}&\cellcolor{gray!12}{55.3}&\cellcolor{gray!12}65.9     \\ \midrule
7 &     &     &     &  \checkmark   &    66.0 {(+3.5)}&72.6&46.0&59.3      \\
\bottomrule
\end{tabular}
\end{center}
\caption{Ablations on VOC 15+5 setting. DLR denotes Decoupled Localization and Recognition. SRD is Semantic-guided Recognition Decoder, DCF means Duplex Classification Fusion. HKD is Hybrid Knowledge Distillation.}
\label{table:ablation-component}
\end{table}
As shown in Table~\ref{table:coco-two-step}, in 70+10 setting, DCA achieves notable improvements of 0.9\% in AP and 1.2\% in AP50 even though the best method stores exemplars to match the training set distribution. When adding more new classes, DCA consistently obtains best in both AP and AP50. 
Notably, improvements become more apparent as the number of categories in the initial stage increases.

\textbf{Multi-step incremental settings.}
Table~\ref{table:coco-20-2} and Table~\ref{table:coco-10-4} show the results under a more demanding setting, where multiple incremental steps are performed to learn new classes. Remarkably, without ER to mitigate forgetting, the performance of existing methods declines drastically. For example, the AP50 of ERD decreased from 66.3\% in the first stage to 31.8\% after learning all classes in four-step setting, nearing 50\% drop. DCA consistently maintains high performance, with AP50 ranging from 68.9\% to 54.1\% in two-step setting and 68.9\% to 49.6\% in four-step setting. DCA exhibits significantly smaller AbsGap, which quantifies the absolute gap to joint training, and RelGap measures the relative gap, indicating that our performance is approaching the upper bound. All results confirm that semantics-guided decoding framework is a powerful and robust incremental detection pipeline.
Meanwhile, unlike architecture-based methods, DCA does not increase many network parameters except for some projection layers for dimensional alignment.

\begin{table}[t]
\small
\setlength\tabcolsep{0.8mm}
	\begin{center}
			\begin{tabular}{c|ccc|cccc}
				\toprule
Row&$ \mathcal{L}^{{kd} } _{{cls}}$ & $\mathcal{L}^{{kd} } _{{vis}}$ & $\mathcal{L}^{{kd} } _{{proj}}$ & All  & Old  & New  & Avg. \\ \midrule
  1&\checkmark  &     &     & 68.8 & 72.7 & 57.2 & 64.9 \\
  2&\checkmark  & \checkmark    &     & 70.5 & 75.0 & 57.1 & 64.6 \\
  3&\checkmark  &     &  \checkmark   & 69.0 & 73.3 & 56.4 & 64.9 \\
  \cellcolor{gray!12}4&\cellcolor{gray!12}\checkmark  &  \cellcolor{gray!12}\checkmark   &  \cellcolor{gray!12}\checkmark   & \cellcolor{gray!12}\textbf{71.2} & \cellcolor{gray!12}76.5 & \cellcolor{gray!12}55.3 & \cellcolor{gray!12}65.9 \\ \bottomrule
\end{tabular}
\end{center}
\caption{Ablation results on Hybrid Knowledge Distillation.}
\label{analysis-hkd}
\end{table}
\subsection{Ablation Studies}
\noindent {\textbf{Component ablations.}}
As shown in Table~\ref{table:ablation-component}, fine-tuning with pseudo-labeling gets the lowest result 62.5\%. Progressive fusions of proposed modules enjoy consistent performance gains. DLR decouples localization and recognition features and reduces the nutual influence of forgetting, while semantic guidance can be precisely applied to the recognition features, maximizing its efficacy. Further combined with HKD, our complete framework surpasses the baseline by up to 8.7\%, while using only HKD results in 66.0\% mAP. These results validate that using semantic guidance during decoding to eliminate reliance on old data is effective and these components are complementary to each other.

\textbf{Analysis of HKD.} 
Results in Table~\ref{analysis-hkd} indicate the critical role of feature regularization $ \mathcal{L}^{{kd} } _{{cls}}$, $\mathcal{L}^{kd}_{vis}$ and $\mathcal{L}^{kd}_{proj}$, verifying that recognition forgetting is significantly influenced by feature drift. Combining with other modules, the forgetting issues in IOD are further mitigated.

\begin{figure}[t]
	\centering
\includegraphics[width=0.7\linewidth]{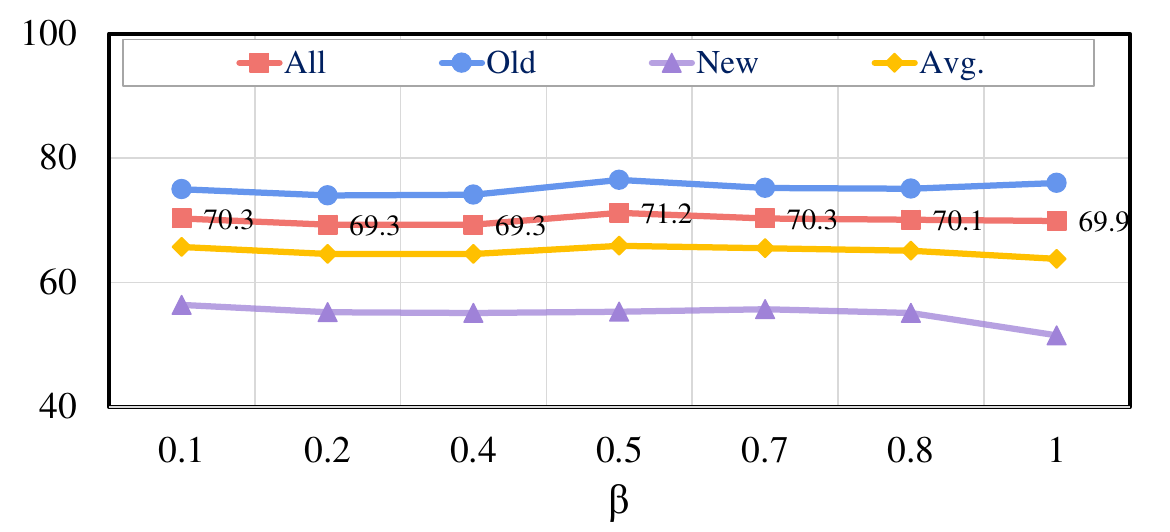}
\caption{Impact analysis about the balance weight. $\beta$ controls the importance of probabilities of the linear classifier.}
\label{fig:stability}
\end{figure}

\textbf{Analysis of the balance weight.} $\beta$ controls the relative importance of the standard linear head and our introduced semantic head in duplex classifier. $\beta$ is set to $\left \{ 0.1,0.2,\dots,1.0 \right \} $. All experiments are run in the 15+5 setting on VOC benchmark, and results are reported in Figure~\ref{fig:stability}. When $\beta=$ 0.5, DCA establishes the best performance on the All metric equally considering all classes, and the Avg. metric equally weighting new and old classes attains the highest. Although duplex classifier fusion increases inference time by 1.6\% (from 0.0687 to 0.0698 s/sample), it is negligible compared to the performance improvement.

\textbf{Robustness to different language models.}
We supplement results of using other language models to obtain class semantic features. In 15+5 setting, results on CLIP, BERT-M~\cite{kenton2019bert} and BERT-S are 71.2\%, 71.0\% and 70.5\% respectively, both higher than 70.2\% of exemplar-free IOD. Our method focuses on leveraging semantic space to boost IOD and is not restricted to the utilization of CLIP text encoder, which offers a viable route to pre-trained semantic spaces.

\section{Conclusions}
\noindent In this paper, we uncover the forgetting imbalance between localization and recognition in transformer-based IOD and propose a Divide-and-Conquer Amnesia (DCA) strategy to effectively mitigate catastrophic forgetting. DCA restructures transformer-based IOD into a localization-then-recognition process, which isolates less-forgetting localization features from the coupled features, thereby leaving the fragile recognition forgetting to be conquered. To reduce recognition feature drift, we leverage semantic knowledge to establish a consistent optimization target throughout the incremental process. For future works, DCA offers a fresh perspective for tackling the challenges of transformer-based IOD, promising significant advancements in the field.
Extensive experiments demonstrate that DCA achieves state-of-the-art, with the advantage of exemplar-free overhead.

\section{Acknowledgments}
\noindent This work is supported by the National Natural Science Foundation of China (Grant NO 62406318, 62376266, 62076195, 62376070), Key Laboratory of Ethnic Language Intelligent Analysis and Security Governance of MOE, Minzu University of China, Beijing, China, and by the Key Research Program of Frontier Sciences, CAS (Grant NO ZDBS-LY-7024).

\bibliography{main}

\end{document}